\documentclass[letterpaper, 10 pt, conference]{ieeeconf}  % Comment this line out if you need a4paper

\IEEEoverridecommandlockouts                              % This command is only needed if 
\usepackage{graphicx}
\usepackage{svg}

\title{\LARGE \bf
Serverless Architecture for Service Robot Management System
}

\author{Kenji Nishimiya$^{1}$ and Yuta Imai$^{2}$% <-this % stops a space
\thanks{$^{1}$Honda R\&D Co., Ltd. 8-1 Honcho Wako-shi Saitama, Japan
        {\tt\small kenji\_nishimiya@jp.honda}}%
\thanks{$^{2}$Soracom Inc, 8F Sankaido Bldg 1-9-13, Akasaka Minato Tokyo, Japan
        {\tt\small imai@soracom.jp}}%
}

\begin{document}

\maketitle
\thispagestyle{empty}
\pagestyle{empty}

%%%%%%%%%%%%%%%%%%%%%%%%%%%%%%%%%%%%%%%%%%%%%%%%%%%%%%%%%%%%%%%%%%%%%%%%%%%%%%%%
\begin{abstract}

We have developed service robot management system to facilitate effective collaboration between multiple units and types of robots in operation. This system is implemented by serverless architecture on cloud and using cellular based IoT communication. So it has not only usual cloud system advantage that it is not necessary to prepare dedicated server and network equipment, but it reduces management efforts of servers. We have tested the system with robots in a public facility, and successfully confirmed its performance and functionality.

\end{abstract}

%%%%%%%%%%%%%%%%%%%%%%%%%%%%%%%%%%%%%%%%%%%%%%%%%%%%%%%%%%%%%%%%%%%%%%%%%%%%%%%%
%% Section I
\section{INTRODUCTION}

\begin{figure}[thpb]
      \centering
      \includegraphics[width=\linewidth]{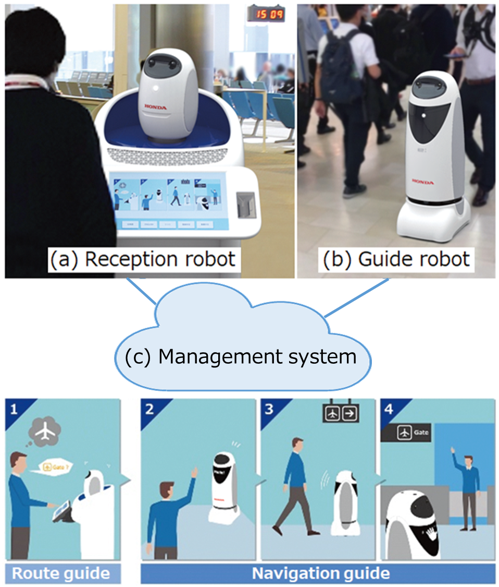}
      \caption{Stationary reception robots in charge of the initial user interaction and mobile guide robots, executing the guiding task, efficiently collaborate under our newly designed management system.}
      \label{scenario}
   \end{figure}

Honda has been engaged in the development of robotics technology, as exemplified by ASIMO, with the aim of realizing “a robot that coexists and cooperates with people and performs useful functions in society \cite{asimo1}\cite{asimo2}. As one such initiative, the guide robot system was developed for robots moving in a mixed environment with people. Guide services by robots are provided to general guests for the purpose of exploring challenges involved in robot implementation in society. The guide robot system is configured with two types of robot, the reception robot and the guide robot (see Fig. \ref{scenario} (a)(b)).

\begin{figure}[thpb]
      \centering
      \includegraphics[width=\linewidth]{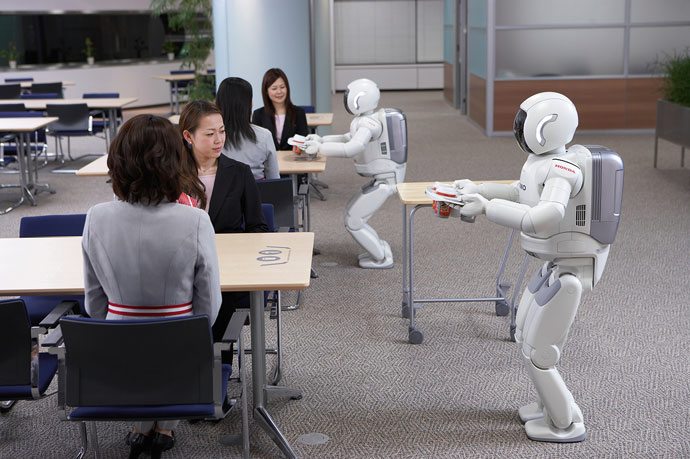}
      \caption{Multiple ASIMO serves drinks to the guests. The cooperative work function enables tasks to be performed cooperatively. A server tracks task status of each ASIMO and distributes tasks to them optimally.}
      \label{asimo}
   \end{figure}

The biggest difference from the robots we have developed so far is that different types of robots collaborate together. Until then, the focus has been on motion planning and control technology for motion generation such as bipedal walking and manipulation, and recognition technology aimed at communication with people. Then, the ability of robots itself became more sophisticated. However, it was assumed that it would operate independently. But in some cases, a demonstration experiment was conducted in which multiple robots were operated. For example, ASIMO provided reception information service and drink service (see Fig. \ref{asimo}) \cite{asimo3}. At this time, we prepared a dedicated server. The management server was sending commands to multiple ASIMO via LAN.

Returning to the topic of this development, the guide robot management system (management system, see Fig. \ref{scenario} (c)) was developed together to allow the network of robots to share information and coordinate between multiple units. The system was implemented with {\it serverless architecture on cloud} and using {\it cellular based IoT communication}.
Serverless architecture is system configuration without any servers, but this system has been built with several cloud native functions that are coordinated together. So there are not only usual cloud system advantages not to have to prepare dedicated servers, but it reduces management efforts of servers for example launch and shutdown, update security patches. In addition, It can adjust the performance of each function, so the performance can be expanded more flexibly.
On the other hand, the IoT (the internet of things) communication can sophisticate the coordination between various products equipped with sensors. IoT has helped accelerate automation in manufacturing, maintenance, etc., with its unique advantage of the light-weight protocol. However, applying IoT to mobile, i.e. wireless, robots requires bi-directional communication under a stable network.

We have reported overview and basic procedure of the guide robot management system already \cite{guide}. This paper elaborates upon technical aspects of the aforementioned serverless architecture and our experiment with a new data communication method.

%% Section II
\section{Related Work}

Network robots have delivered promising results in many sectors. Particularly in the industrial sector, they have shown a beyond-human capability in managements factories and warehouses. As well, there are successful applications of remote-controlled robots in surgical operations. 

The term {\it Cloud Robotics} was coined by James J. Kuffner in 2010 \cite{cloud_robotics}. He described its potential. In recent years, with the advancement of cloud computing, a large amount of research about its implementation with robots has been conducted \cite{1}. There are two main objectives for using cloud-computing with robots. Firstly, we aimed to economize the computing capacity by allocating it in a cloud rather than in each robot. This method has been implemented in numerous studies over years. For example, the idea {\it Remote-brained robots} was described by M. Inaba in 1993 \cite{remote_brain}.
Recently adoption of machine learning to increase the processing capability of data and operations on the cloud has also undergone trials \cite{3}. 
Secondly, sharing of information between multiple robots would enable them to conduct tasks that would otherwise be impossible.
{\it DAvinCi} project in 2010 demonstrated the application for SLAM collecting messages from multiple robots. They proposed the cloud architecture to offload data and computation using Hadoop cluster, moreover these robots could exchange useful data \cite{davinci}. Sharing data and learn example is the {\it RoboEarth project} \cite{4}. An extensive open-source network and knowledge database are accessible.

When building a system of network connected robots, including both cloud-to-robot and robot-to-robot communication channel, some technical restrictions such as communication data size and network latency, would have to be considered. In this way, an appropriate system configuration is necessary for each use-case \cite{2}.

On the other hand, {\it Rapyuta} a generic cloud robotics platform for commercial uses has been released \cite{5}. ROS (Robot Operating System) is a widely used middleware in robotics research \cite{ros} that operates on cloud servers that are connected to the robots via WebSocket. They demonstrated cloud based mapping as an example application.

Most of the related research so far has been to set up a server or servers on cloud and connect it to robots.

%% Section III
\section{Overview of the Guide Robot}
\subsection{Reception Robot and Guide Robot}

The guide robot system provides two types of service.
One is the route guide and the other is the navigation guide. These services are assigned to reception robots and guide robots respectively, but their execution is coordinated. When guests specify their destination either by voice or touch screen terminal input, the reception robot indicates the route to the destination (route guide function). Then, the guide robot directs the way to their destination, while accompanying them along the journey (guide function).

In order to guide guests, guide robot has autonomous navigation function and facial recognition function. Facial recognition identifies the guests using the camera installed at the rear of guide robot and recognizing their relational position \cite{face}. It is used to make sure the robot speed is suitable for the guests and notice separation from the guests. To this end, the two types of robots share the facial profile of the guests in addition to the destination’s coordinate.

\subsection{Management System}
The management system has two main roles to play.

The first role is to manage fleet of guide robots and these tasks so that the system can respond to guide requests from reception robots. Guide requests with the guest’s desired destination and facial feature data which is acquired by the reception robot. Based on operational status, remaining battery charge and accumulative mileage, the system selects a particular robot to allocate a task. In most cases, during operation, the robot with the shortest mileage accepts the request. It is simply to level the mileage among the robots.
The system sends the data as a set of commands to the guide robot, which receives the data and initiates operation.

The second role is to monitor the operational status of robots and notify the system operator of any malfunctions. During guide operations, the management system constantly monitors the progress of tasks. When a certain task is completed, the next command is sent. The system operator also carries an operation monitoring terminal by which the location and operational status of guide robots can be checked, and when trouble occurs, detailed information regarding the location and nature of the malfunction can be checked through the terminal.

%% Section IV
\section{Overview of the Management System}

\begin{figure}[thpb]
      \centering
      % \includesvg[width=\linewidth,inkscapelatex=false]{figs/management_system.svg}
      \includegraphics[width=\linewidth]{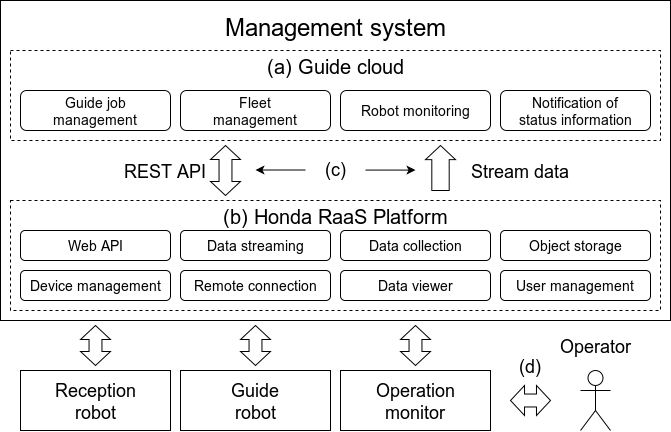}
      \caption{Functional layers of the management system. Honda RaaS provides basic functions for robot operation. Depend on the application, another cloud functions can be expanded like this guide cloud.}
      \label{management_system}
   \end{figure}

The management system is made up of the Honda Robotics as a Service Platform (Honda RaaS) and the guide cloud application (guide cloud). Honda RaaS (see Fig. \ref{management_system} (b)) is a platform for development and operation of robotics services, which provides general purpose features and services such as message transmission among cloud/robots, storage service and device management \cite{raas}.

In contrast, the guide cloud (see Fig. \ref{management_system} (a)) was developed additionally to realize particular functions specific to guide operation. It uses data shared from Honda RaaS as the basis for managing guide jobs and monitoring operational status of the robots. Honda RaaS and the guide cloud are linked by data streaming and REST API (see Fig. \ref{management_system} (c)). Reception robots, guide robots, and operation monitoring terminals are connected through the Internet using IoT and other such frame of communication. 

   \begin{figure*}[thpb]
      \centering
      \includegraphics[width=\linewidth]{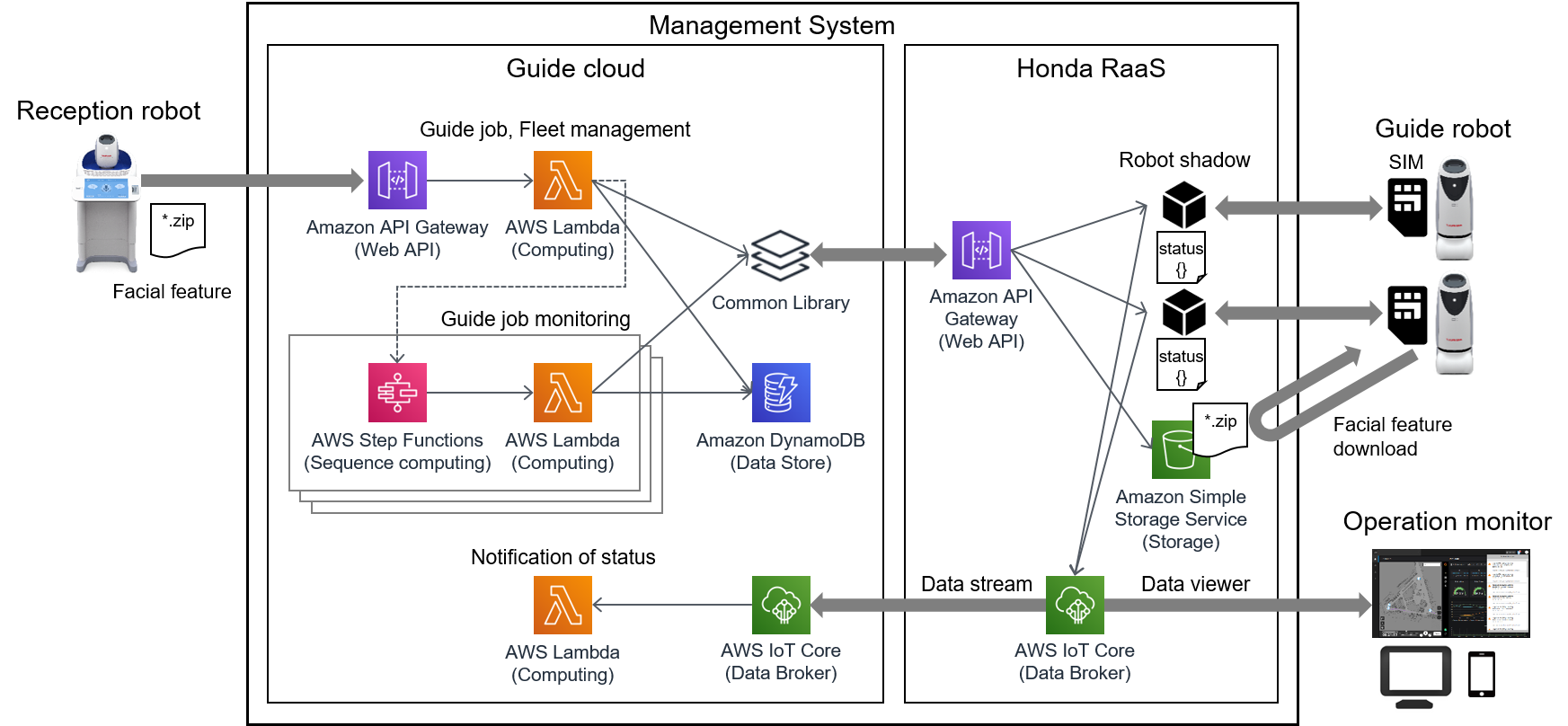}
      \caption{Serverless architecture of the management system: Square blocks are cloud native functions, only executed when the operation is demanded. Data will be transferred between these functions.}
      \label{serverless}
   \end{figure*}

%% Section V
\section{Management System Architecture and Main Functions}

The management system takes advantage of serverless architecture and utilizing cloud native services (see Fig. \ref{serverless}). Serverless enables to build applications without preparing servers \cite{aws}. For example, we use AWS Lambda for computing, Amazon DynamoDB for database, Amazon S3 for storage and AWS Step Functions for task management for robots. Serverless architecture, combined with online computing services, has three advantages. (1) It does not require new hardware. (2) The scale and other features of the service are flexible when faced with varying demand. (3) Up-to-date security standard is maintained. For communication part, we use Honda RaaS as we discussed before, it provides generic data communication and robot identification services by integrating with the cellular based IoT communication. By using this, (1) we eliminated constraints of operational range and (2) leveraged SIM (Subscriber Identity Module) card as a credential for robots so that application developers do not have to manage it. Thus, we can dedicate on application development itself.

\subsection{Robot shadow and guide job monitoring}

A request for guide service received by a reception robot is then processed by the management system which allocates the task to a guide robot. To do this, the system keeps track of the guide robot's location, battery and operational status. The system collects from a robot at various time interval status information to save the communication capacity. Such an interval is set based on the rate at which the type of status changes. For example, the location would refresh every 2.0 seconds while the battery level, every 10.0 seconds. Even if the communication with a robot is somehow not available, the system can utilize the last recorded status of the robot, which we refer to as 'Robot Shadow'.

\begin{figure}[thpb]
      \centering
      \includegraphics[scale=0.5]{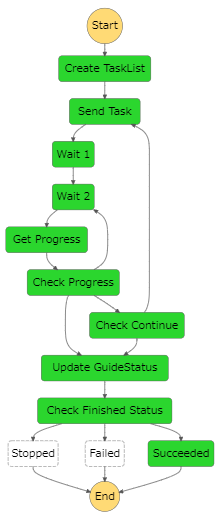}
      \caption{Guide job monitoring sequence: This sequence can be launched for each guide jobs in parallel, and it monitors task status periodically.}
      \label{step_functions}
   \end{figure}

The management system launches a guide job monitoring sequence in Fig. \ref{step_functions} as soon as the robot begins guiding guests. When the task is completed, the system then sends information for the next task. This monitoring sequence is implemented and run on function as a service (here, we use AWS Step Functions), it is spun up on guide job and spun down when the job has been finished. Thus we are free from managing this type of scenario server.

\subsection{File transfer from the management system to robots}
The facial feature of the guests is obtained with their permission by the reception robot's built-in camera while showing the guests the route guide. The data is then sent to the guide robot via the management system, which uses it to recognize the relative position of the guests when guiding.
The system protects the guests' privacy by encrypting the facial profile as well as by using a signed URL for downloading.

\subsection{Operation Monitor}

\begin{figure}[thpb]
      \centering
      \includegraphics[width=\linewidth]{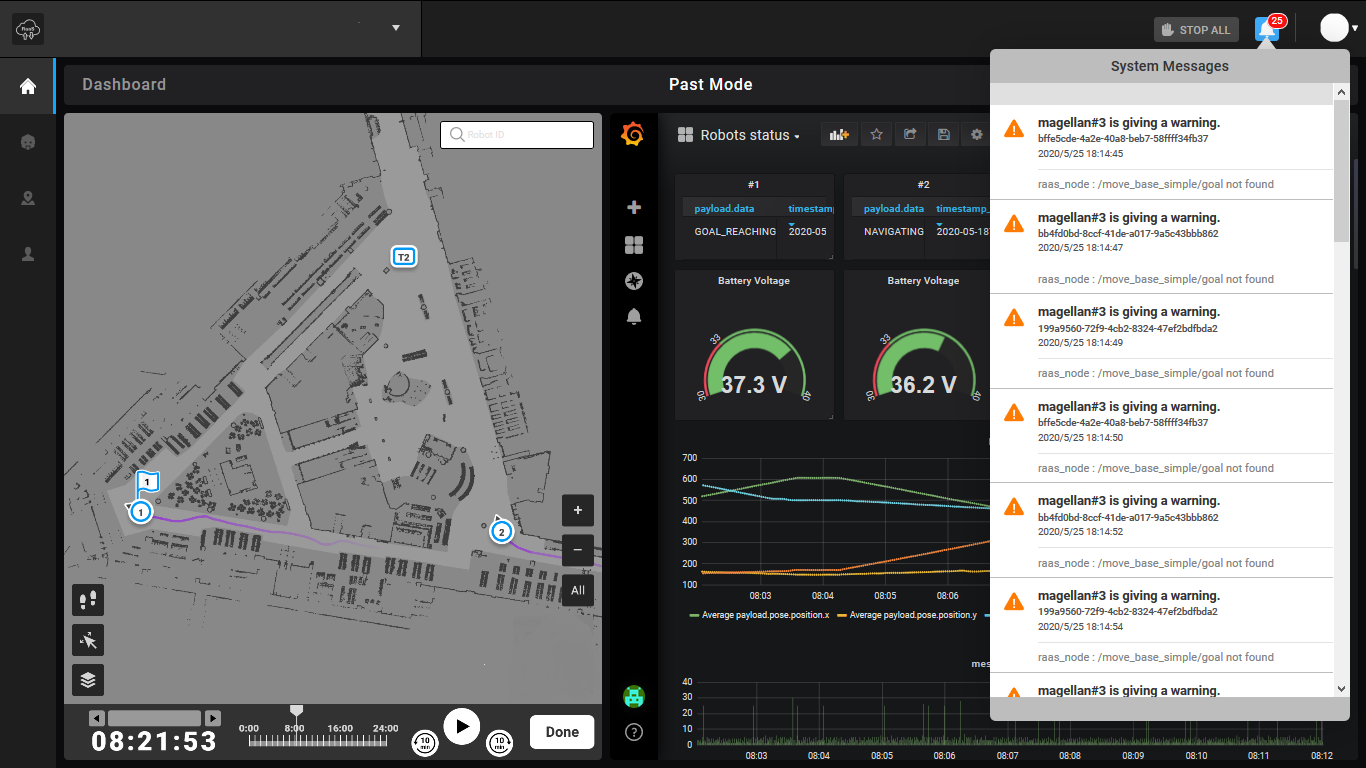}
      \caption{A screenshot of the operation monitor: The device runs a web-browser based application. The position, path and destination of robots are shown on the left-hand side, of which the colour corresponds to the status of each unit; blue for normal, green, working, red, an error, etc. The middle part shows detailed data received from the robots. The operator would receive pop-up notifications on the right-hand side, where past monitoring log data can is collected, which the operator can review using the slider bar.}
      \label{dashboard}
   \end{figure}

The system operator can remotely monitor the status of robots with a device called operation monitoring terminal which displays data from the management system as seen in Fig. \ref{dashboard}. The operator is notified in an even of contingencies or system troubles through monitoring terminal or via email.

%% Section VI
\section{Evaluation}

\subsection{Performance Test}
Researches indicated that people would have a worse impression when the robot takes more than about 2.0 seconds to respond \cite{ui}. Therefore, a timely response to the command is crucial for the user experience.
Particularly, when using a wireless connection, the latency and interruption have to be carefully considered.

\begin{figure*}[thpb]
      \centering
      \includegraphics[width=\linewidth]{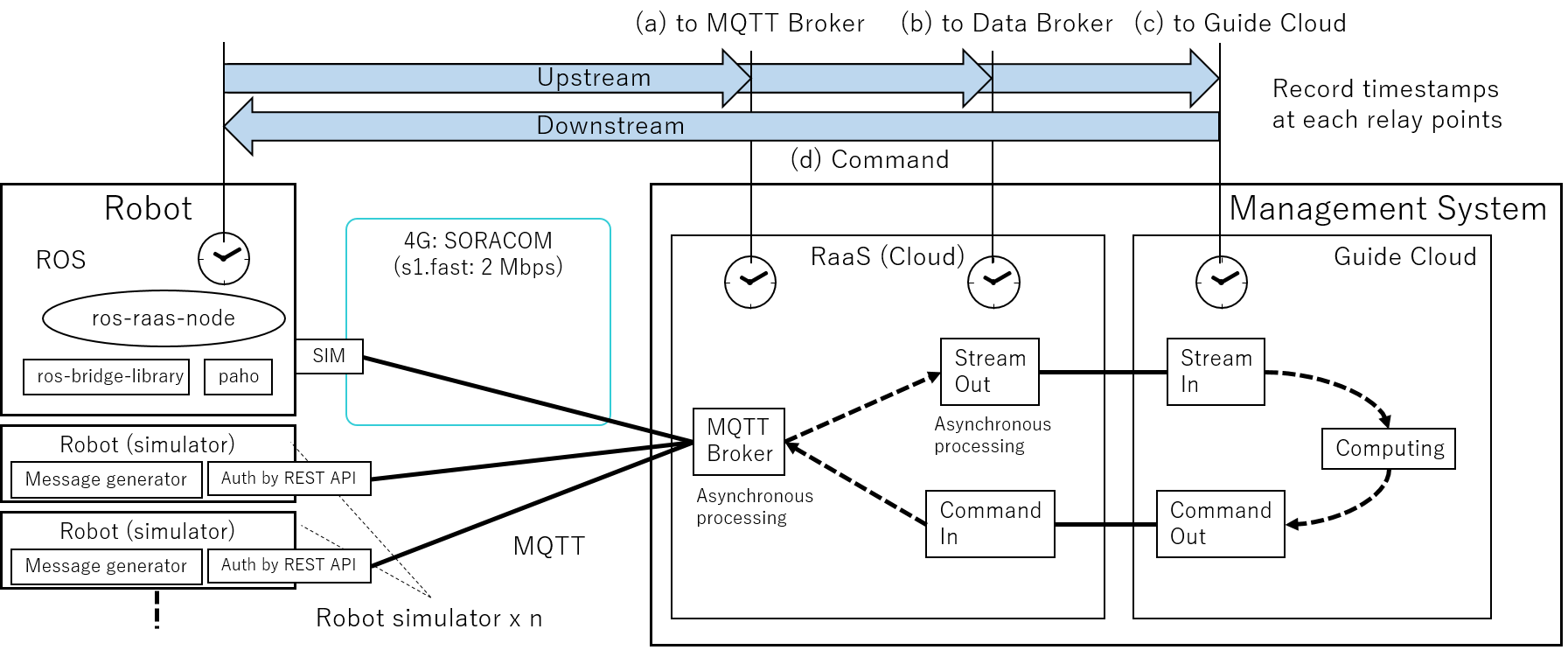}
      \caption{Configuration diagram of data communication performance tests. Upstream and Downstream latency was measured by recording timestamps at each relay points. There are two asynchronous processing on the upstream path to transfer streaming data. To reproduce communication load up to 100 units, simulator which can generated data were used in the test.}
      \label{performance_test}
   \end{figure*}

To this end, we have conducted performance tests for two-way communication between the robots and the management system which uses IoT communication facilitated by the cellular network (4G, SORACOM). The expected turn-around time (upstream + downstream) after a user interaction was under 1.0 second, where upstream signals are data acquisition and downstream, commands. The latency was measured as a gap between the recorded timestamps during the end-to-end communications accompanied by asynchronous data processing in RaaS that are MQTT broker and data stream out (see Fig. \ref{performance_test}).

Our preliminary survey conducted at a real-life facility on the demand for guiding service showed that no more than 100 requests for such service at any given time are to be expected. Therefore in simulated communication load was of up to 100 units. Since we found it unfeasible to prepare many robots for the experiment, we used a simulator so that, RaaS can register virtual robots as if we had actual robots. The accordingly-generated logging data of virtual roots replicate the streaming processing load on the management system, even devoid of wireless communications.

\begin{figure}[thpb]
      \centering
      % \includesvg[width=\linewidth,inkscapelatex=false]{figs/plot_avg.svg}
      \includegraphics[width=\linewidth]{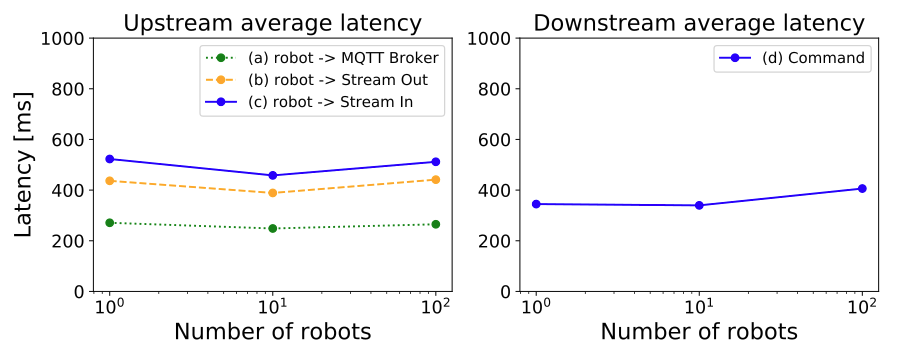}
      %\framebox{Turn around time: (c) + (d) \textless 1.0}
      \caption{Average latency between robots and the management system with varying number of connected robots 1, 10 and 100. Upstream latency is average value of continuous data acquisition for 10 minutes which contains two relay points on the path and end point. Downstream latency is average value of 10 commands while collecting data. Target turn around time (c) + (d) is within 1.0 second and could be cleared under those number of robots.}
      \label{plot_avg}
   \end{figure}

\begin{figure}[thpb]
      \centering
      % \includesvg[width=\linewidth,inkscapelatex=false]{figs/plot_data.svg}
      \includegraphics[width=\linewidth]{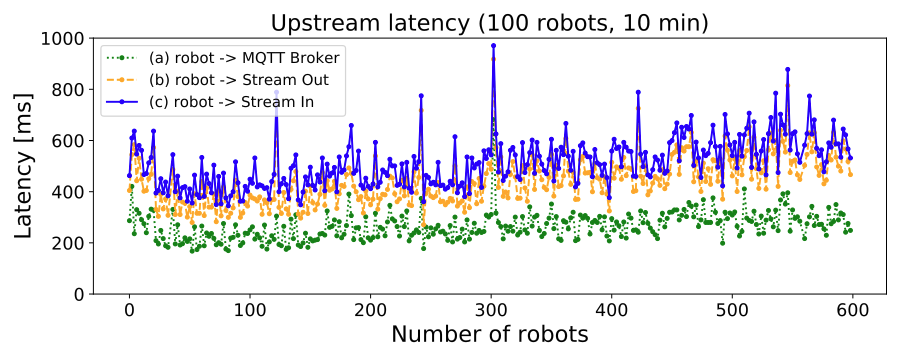}
      %\framebox{Stable but sometimes larger delay occurred.}
      \caption{10 minutes time-series latency when 100 robots (using simulator) connected. There is no interruption and the performance was stable. Some larger delay, however, were also recorded in some occasions; 5-10 times during a 10 minute trial.}
      \label{plot_data}
   \end{figure}

Average latency of 400 - 550 ms without interruption for upstream and 300 - 400 ms for downstream, were observed (see Fig. \ref{plot_avg}, \ref{plot_data}), adding up to less than 1.0 second for back and forth communication. As the data was increased the load balancer was activated to expand computer resources automatically. Moreover, we have confirmed that the system can handle a large number of robots. However, we observed larger delay in some specific communication environment (see Fig. \ref{plot_data}). Such unpredictable delays occurred at a data streaming process from RaaS to guide cloud. RaaS utilizes AWS IoT Core to serialize and transfer messages from a number of robots. This cloud service provides MQTT interfaces that make it easier to implement scalable and reliable data processing \cite{iotcore}, but, it does not provide guarantee for latency. For this reason, we saw such unpredictable delay. Therefore designing a sequence to retain user's attention will be necessary to further improve the user experience. Nonetheless, the overall results were positive, as we confirmed the capability of the system to monitor robots' status and execute the fleet algorithm.

\subsection{Function Test}

Confirmation of service envisioning guide services and using actual equipment was carried out at a certain public indoor facility which extends as far as 500 meters in each direction and include many tenants run businesses. New visitors to the facility may find it challenging to figure out what direction to take to reach their destination. 
We confirmed that the management system functions in the following manner during the testing.

\begin{itemize}

\item One appropriate robot would be selected from the guide robots in accordance with the conditions for operational status and remaining battery charge.
\item The transmission of destination and facial feature data to the guide robot with appropriate timing and other such matters were confirmed. The guide robot recognized the guests correctly and performed the configured tasks of pickup, guidance, and return.
\item As it moved throughout the entire facility, the guide robot has successfully transmitted the data needed to update its movement status on the system. We, therefore, demonstrated that such a method of communication is comparable to the performance test in the previous section. Hence, the robot-monitoring and the transmission of commands could be carried out in a stable manner using cellular-based IoT communication, quod erat demonstrandum.
\end{itemize}

\begin{figure*}[thpb]
      \centering
      \includegraphics[width=\linewidth]{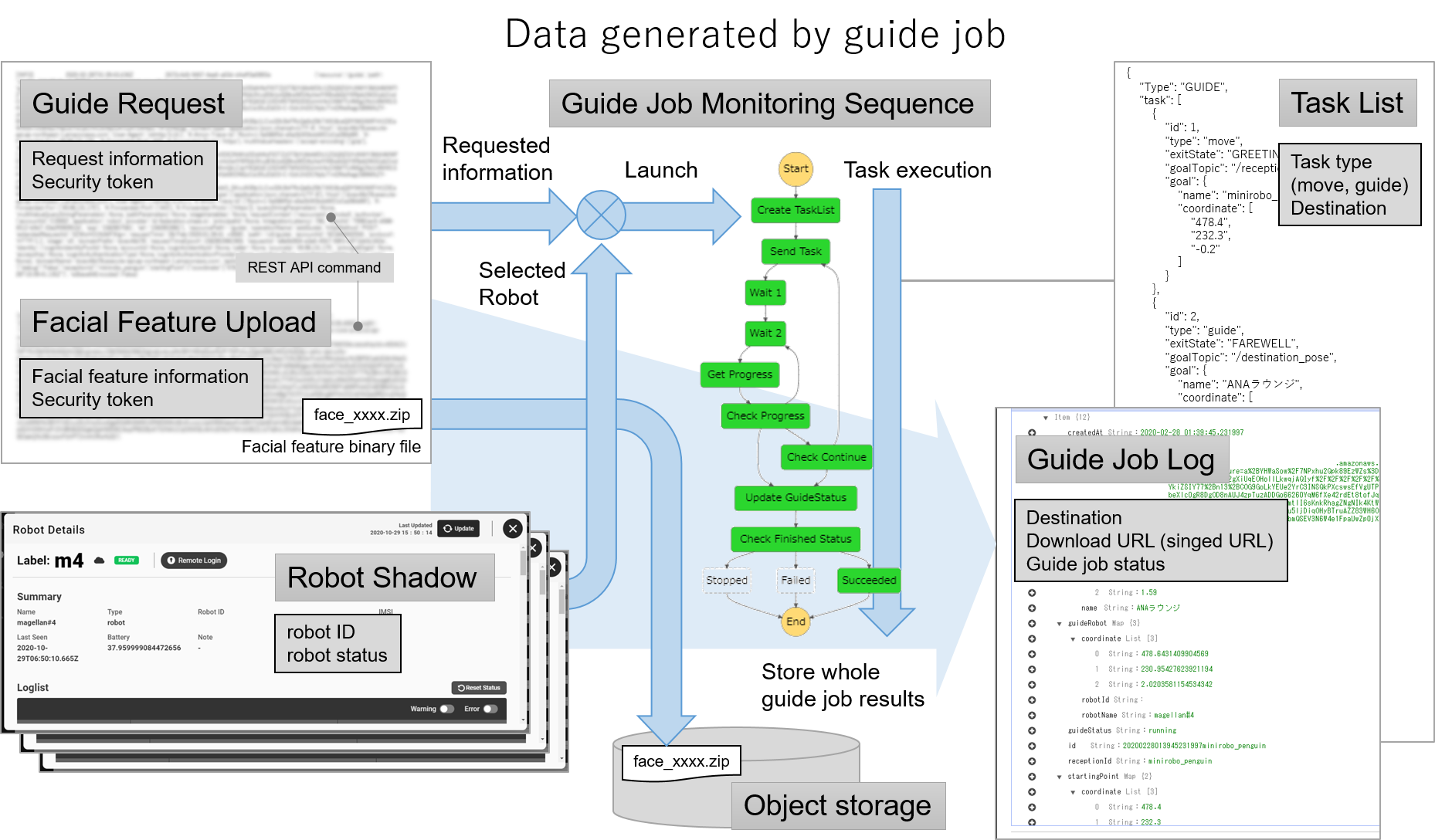}
      \caption{An example of data generated as a result of a guide job, each section corresponding to different task; from the request, information, for example, destination was transferred to the next procedure. Guide job monitoring sequence generated log with each step which would be stored for later analysis. }
      \label{data_flow}
   \end{figure*}

Between the testing, we have checked the log data of each function such as received API commands, robot selection algorithm determined by the status of active robots, contents of the task list and the behavior of guide job monitoring sequence (see Fig. \ref{data_flow}). Since most of the functions are linked by multiple services, detailed log checks must be done to build the data flow correctly. The cloud logging function automatically records logs without implementing code for log generation, which helped us to perform detailed analysis.

Eventually, we confirmed that the system performs guide job management and fleet management according to the design and that it is capable of monitoring the operational status of the robots using IoT communication.

%% Section VII
\section{Conclusion}

We constructed a system that consists of serverless architecture on cloud and cellular based IoT communication, enabling the management of multiple units and types of service robots.
The system successfully performed the assigned task during tests at a real public facility simulating delivery of the guide service.
The potential of integration between robots and cloud technology has been pointed out for a long time. We confirmed possibilities specifically in using a serverless architecture 

\begin{itemize}
\item Easier addition of new functions
\item Expansion of computing capacity
\item Reduction of operating efforts
\newline
\end{itemize}

%\clearpage

\end{document}